\documentclass[11pt]{article}
\usepackage{coling2020}
\usepackage{times}
\usepackage{url}
\usepackage{latexsym}

\usepackage{epsfig}
\usepackage{amsmath}
\usepackage{amssymb}
\usepackage{graphicx}
\usepackage{epstopdf}
\usepackage{wrapfig}
\usepackage{subfigure}
\usepackage{caption}
\usepackage{verbatim}
\usepackage{algorithm}
\usepackage{algpseudocode}
\usepackage{multirow}
\usepackage{booktabs}
\usepackage{amssymb}
\usepackage{bm}
\usepackage{footnote}

\colingfinalcopy % Uncomment this line for the final submission

\title{LightMBERT: A Simple Yet Effective Method for \\Multilingual BERT Distillation}

\author{Xiaoqi Jiao$^{1}$\thanks{\hspace{0.5pt} This work is done when Xiaoqi Jiao is an intern at Huawei Noah's Ark Lab.} , Yichun Yin$^{2}$, Lifeng Shang$^{2}$, Xin Jiang$^{2}$\\
	\textbf{Xiao Chen$^{2}$, Linlin Li$^{3}$, Fang Wang$^{1}$ and Qun Liu$^{2}$}\\
	$^{1}$Huazhong University of Science and Technology\\
	$^{2}$Huawei Noah's Ark Lab, $^{3}$Huawei Technologies Co., Ltd.\\
	\texttt{\{jiaoxiaoqi,wangfang\}@hust.edu.cn}\\
	\texttt{\{yinyichun,shang.lifeng,jiang.xin\}@huawei.com}\\
	\texttt{\{chen.xiao2,lynn.lilinlin,qun.liu\}@huawei.com}
}

\date{}

\begin{document}
\maketitle
\begin{abstract}
	The multilingual pre-trained language models~(e.g, mBERT, XLM and XLM-R) have shown impressive performance on cross-lingual natural language understanding tasks. However, these models are computationally intensive and difficult to be deployed on resource-restricted devices. In this paper, we propose a simple yet effective distillation method~(LightMBERT) for transferring the cross-lingual generalization ability of the multilingual BERT to a small student model. The experiment results empirically demonstrate the efficiency and effectiveness of LightMBERT, which is significantly better than the baselines and performs comparable to the teacher mBERT. 
	
	%In addition, ablation studies are also conducted to explain the contribution of each procedure of our method.
\end{abstract}

%\begin{abstract}
%  The multilingual BERT~(mBERT), pre-trained on Wikipedia corpus from 104 languages, has shown impressive performance on the cross-lingual natural language inference benchmark~(XNLI). In this paper, we explore the possibility of transferring the cross-lingual generalization of mBERT to a small student by knowledge distillation~(KD). Experiments empirically demonstrate that the proposed Effective and Efficient Multilingual Distillation~(E2MD) method is significantly better than the baselines with only 50\% training steps, and achieve comparable results to the teacher mBERT. In addition, ablation studies are also conducted to explain the contribution of each procedure of our method.
%\end{abstract}

\section{Introduction}
\label{intro}

%
% The following footnote without marker is needed for the camera-ready
% version of the paper.
% Comment out the instructions (first text) and uncomment the 8 lines
% under "final paper" for your variant of English.
% 
%\blfootnote{
    %
    % for review submission
    %
    %\hspace{-0.65cm}  % space normally used by the marker
    %Place licence statement here for the camera-ready version.
    %
    % % final paper: en-uk version 
    %
    % \hspace{-0.65cm}  % space normally used by the marker
    % This work is licensed under a Creative Commons 
    % Attribution 4.0 International Licence.
    % Licence details:
    % \url{http://creativecommons.org/licenses/by/4.0/}.
    % 
    % % final paper: en-us version 
    %
    % \hspace{-0.65cm}  % space normally used by the marker
    % This work is licensed under a Creative Commons 
    % Attribution 4.0 International License.
    % License details:
    % \url{http://creativecommons.org/licenses/by/4.0/}.
%}

Multilingual pre-trained language models~(PLMs), such as mBERT~\cite{devlin2019bert} and XLM~\cite{conneau2019cross}, have been shown to be effective on a variety of cross-lingual benchmarks~\cite{conneau2018xnli,artetxe2019cross,hu2020xtreme}. Moreover, XLM-R~\cite{conneau2019unsupervised} further demonstrates that it is possible to have a single large model for all languages, without sacrificing per-language performance. But, in practice, it is challenging to deploy these large multilingual PLMs on resource-limited devices because of the latency and memory requirements.

Knowledge distillation~(KD)~\cite{hinton2015distilling}, one of the model compression techniques, has been successfully used for compressing monolingual PLMs~\cite{tang2019distilling,sun2019patient,sanh2019distilbert,tsai2019small,jiao2019tinybert,turc2019well,sunmobilebert,wang2020minilm}. It transfers the knowledge embedded in a large teacher PLM to a small student model by making the student model mimic the behaviors of the teacher. In this paper, we focus on the cross-lingual scenario where there is no task training data in the target languages, which is different from previous works~\cite{tsai2019small,mukherjee2020xtremedistil}. Specifically, we take mBERT as an example and investigate how to effectively and efficiently distill the cross-lingual generalization ability of it into a transformer~\cite{vaswani2017attention} based student.

To achieve efficient multilingual distillation, we first initialize the student with the bottom layers of mBERT, which allows the student model to directly inherit some knowledge of mBERT at the beginning. Then, we freeze the inherited embeddings during the distillation process since they are shown to be important for the cross-lingual generalization ability of mBERT~\cite{pires2019multilingual,wu2019beto}. Last, we perform the {\it top transformer-layer distillation} on unsupervised corpus in multiple languages.

Experiments on the cross-lingual natural language inference benchmark XNLI~\cite{conneau2018xnli} demonstrate that our distillation method significantly outperforms the baselines and achieves comparable results to the teacher mBERT.

\section{Method}

\begin{figure}
	\centering
	\includegraphics[scale=0.7]{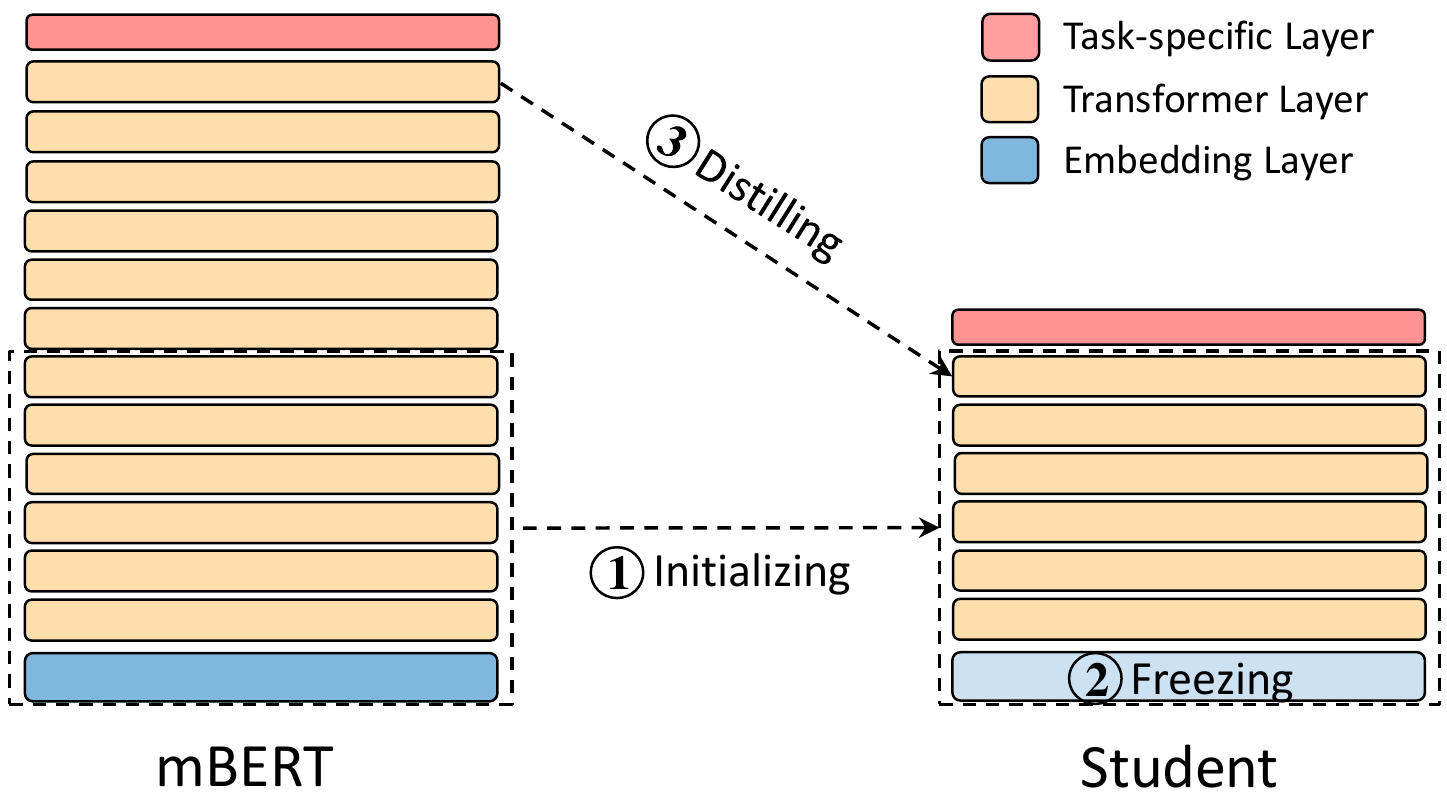}
	\caption{The Diagram of LightMBERT. Lighter blue block represent that they are frozen during the distillation process.}
	\label{figure:distill_mBERT}
\end{figure}

In this section, we detail our multilingual knowledge distillation method, which includes three stages: 1) Initializing the embedding and transformer layers of a student model with the embedding and bottom transformer layers of the teacher mBERT; 2) Freezing the embedding layer of the student model; 3) Performing the top transformer-layer distillation on large-scale unsupervised corpus in different languages. The diagram of our proposed distillation method is shown in Figure~\ref{figure:distill_mBERT}.

\paragraph{Initialization} Performing task-agnostic distillation of PLMs is time-consuming since the large-scale corpus for pre-training these PLMs are also used for transferring the knowledge embedded in them. For multilingual PLMs, it becomes unacceptable because of the much more training corpus~(e.g. Wikipedia from 104 languages are used in the pre-training stage of the multilingual BERT). Our solution to reduce the training time of multilingual distillation is to inherit some knowledge from the teacher model at the beginning. Inspired by the pruning work of~\newcite{sajjad2020poor}, in which they explored the straight-forward strategies of dropping some layers of the PLMs~(e.g. BERT, RoBERTa), and found the top-layer dropping strategy works consistently well for all models. Therefore, we initialize our student model with the bottom layers from the teacher mBERT. 

\paragraph{Freezing the Embeddings} The embeddings, shared across languages, is considered important for the cross-lingual generalization ability in the multilingual BERT~\cite{pires2019multilingual,wu2019beto}. \newcite{Cao2020Multilingual} further show that the embeddings learned by mBERT already somewhat are aligned across languages. By initializing the embeddings of a student by that of mBERT and freezing it from updating, we can ensure the embeddings of the student contains rich multilingual information and also ease the training process.

%The uniform {\it transformer-layer distillation}, introduced by TinyBERT~\cite{jiao2019tinybert}, learns evenly from the transformer layers of the teacher network. Unlike TinyBERT
\paragraph{Top-layer Distillation} We distill the knowledge from the top transformer layer of mBERT into the student instead from every few layers which is adopted by TinyBERT~\cite{jiao2019tinybert}. The knowledge in a transformer layer includes the attention and hidden states. The loss of attention based distillation is: 
\begin{align}
\label{eq:att_loss}
\mathcal{L}_{\text{attn}} = \frac{1}{h}\sum\nolimits^{h}_{i=1} \texttt{MSE}(\bm{A}_i^{S}, \bm{A}_i^{T}),
\end{align}
where $h$ is the number of attention heads, $\bm{A}_i^{S}$ and $\bm{A}_i^{T}$ refer to the attention matrix corresponding to the $i$-th head of student and teacher respectively, and {\tt MSE()} means the {\it mean squared error} loss function. The hidden states knowledge is defined as the output of a transformer layer, and the objective of hidden states based distillation is:
\begin{align}
\label{eq:hid_loss}
\mathcal{L}_{\text{hidn}} = \texttt{MSE}(\bm{H}^{S}, \bm{H}^{T}), 
\end{align}
where $\bm{H}^{S}$ and $\bm{H}^{T}$ refer to the hidden states of student and teacher networks respectively. Finally, the total loss of the transformer-layer distillation is:  
\begin{align}
\label{eq:transformer_layer_loss}
\mathcal{L}_{\text{layer}} = \mathcal{L}_{\text{attn}} + \mathcal{L}_{\text{hidn}}.
\end{align}

\section{Experiments}
In this section, we empirically study the effectiveness and efficiency of our proposed multilingual knowledge distillation method.
\vspace{-0.2cm}
\paragraph{Dataset} We evaluate the distilled small model on the Cross-lingual Natural Language Inference~(XNLI)~\cite{conneau2018xnli} benchmark, which includes ground-truth dev and test sets in 15 languages and a ground-truth English training set. 
\vspace{-0.1cm}
\paragraph{Baselines} We compare LightMBERT with three representative baselines: 1) DistilmBERT, which is the multilingual version of DistilBERT~\cite{sanh2019distilbert}; 2) mBERT\_drop that refers to directly pruning the top transformer layers of mBERT. This is a straight-forward layer pruning strategy proposed by~\newcite{sajjad2020poor}; 3) Multilingual TinyBERT, denoted as TinyMBERT. We use the released code\footnote{\url{https://github.com/huawei-noah/Pretrained-Language-Model/tree/master/TinyBERT}} of TinyBERT~\cite{jiao2019tinybert} to perform knowledge distillation in multiple languages at pre-training stage.  For a fair comparison, the architecture of all the student models is the same as DistilmBERT, which has 6 transformer encoder layers with 768 hidden size.
\vspace{-0.1cm}
\paragraph{Setting} The teacher and tokenizer used in our experiments is the same as mBERT~\cite{devlin2019bert} and we use Wikipedia in multiple languages as the training corpus, which is extracted by the WikiExtractor tool\footnote{\url{https://github.com/attardi/wikiextractor}}. We set the batch size to 256, peak learning rate to 1e-4, and train the 6-layer students with a maximum sequence length of 128 for 400, 000 steps, the typical training steps adopted in monolingual KD setting~\cite{wang2020minilm}. We use linear warmup over the first 40, 000 steps and linear decay. The dropout rate is 0.1. The weight decay is 0.01. In addition, for fine-tuning on XNLI, the hyper-paremeters are fixed as follow: the epoch number, learning rate, batch size and maximum sequence length are 3, 2e-5, 32 and 128, respectively. Still, We freeze the embedding layer of LightMBERT at fine-tuning stage.
\vspace{-0.6cm}
\paragraph{Results} Same with DistilmBERT, we here report zero-shot cross-lingual transfer results on 6 languages, as shown in Table~\ref{tab:main_results}. Zero-shot means that the multilingual models are fine-tuned on the English training set, and then evaluated on other language XNLI test sets. The experiment results demonstrate that: 1) LightMBERT outperforms all the baselines by a margin of at least 2.1\% on average, and achieves comparable results with its teacher mBERT; 2) Surprisingly, mBERT\_drop, obtained by directly pruning the top 6 transformer layers of mBERT, performs better than DistilmBERT. This interesting observation indicates that mBERT\_drop retain a lot of knowledge of mBERT. 3) Note that our method, including initializing a student by mBERT\_drop and continuing to transfer the knowledge from mBERT to the student by the top {transformer-layer distillation} brings extra 3.2\% improvement which confirms it is feasible to re-acquire the lost knowledge, caused by the above mentioned straight-forward pruning method.

\begin{savenotes}
\begin{table}[]
	\centering
	\begin{tabular}{|c|c|c|c|c|c|c|c|}
		\hline
		     \textbf{Model}      & \textbf{English} & \textbf{Spanish} & \textbf{Chinese} & \textbf{German} & \textbf{Arabic} & \textbf{Urdu} & \textbf{AVG} \\ \hline
		MBERT~(Teacher)$\dagger$ &       82.1       &       74.6       &       69.1       &      72.3       &      66.4       &     58.5      &     70.5     \\ \hline \hline
		  DistilmBERT$\dagger$   &       78.2       &       69.1       &       64.0       &      66.3       &      59.1       &     54.7      &     65.2     \\ \hline
		      mBERT\_drop        &       79.1       &       71.8       &       66.5       &      67.7       &      61.2       &     56.1      &     67.1     \\ \hline
		       TinyMBERT         &       80.5       &       72.3       &       67.2       &      68.4       &     63.5        &     57.5      &     68.2     \\ \hline
		   LightMBERT~(Ours)     &       \textbf{81.5}       &       \textbf{74.7}       &       \textbf{69.3}       &      \textbf{72.2}       &      \textbf{65.0}       &     \textbf{59.3}      &     \textbf{70.3}     \\ \hline
	\end{tabular}
	\vspace*{-0.1cm}
	\caption{The zero-shot cross-lingual transfer results on XNLI test sets. All the student models have the same architecture~(6 transformer layers with 768 dimensions) and are task-agnostically distilled from mBERT. $\dagger$~denotes that the results are taken from DistilmBERT~\footnote{\url{https://github.com/huggingface/transformers/tree/master/examples/distillation}}.} 
	\label{tab:main_results}
	\vspace{-0.2cm}
\end{table}
\end{savenotes}

\section{Ablation Studies}
In this section, we conduct experiments to investigate the contributions of the three different parts in our method: 1) Initialization that our student is initialized by the embedding and the bottom transformer layers of the teacher mBERT; 2) Freezing the shared sub-word embeddings during the distillation process and 3) Performing the transformer-layer distillation only at the top transformer layer of the student. 

\begin{figure}[t]
	\centering
	\subfigure[English]{\includegraphics[width=0.4\textwidth]{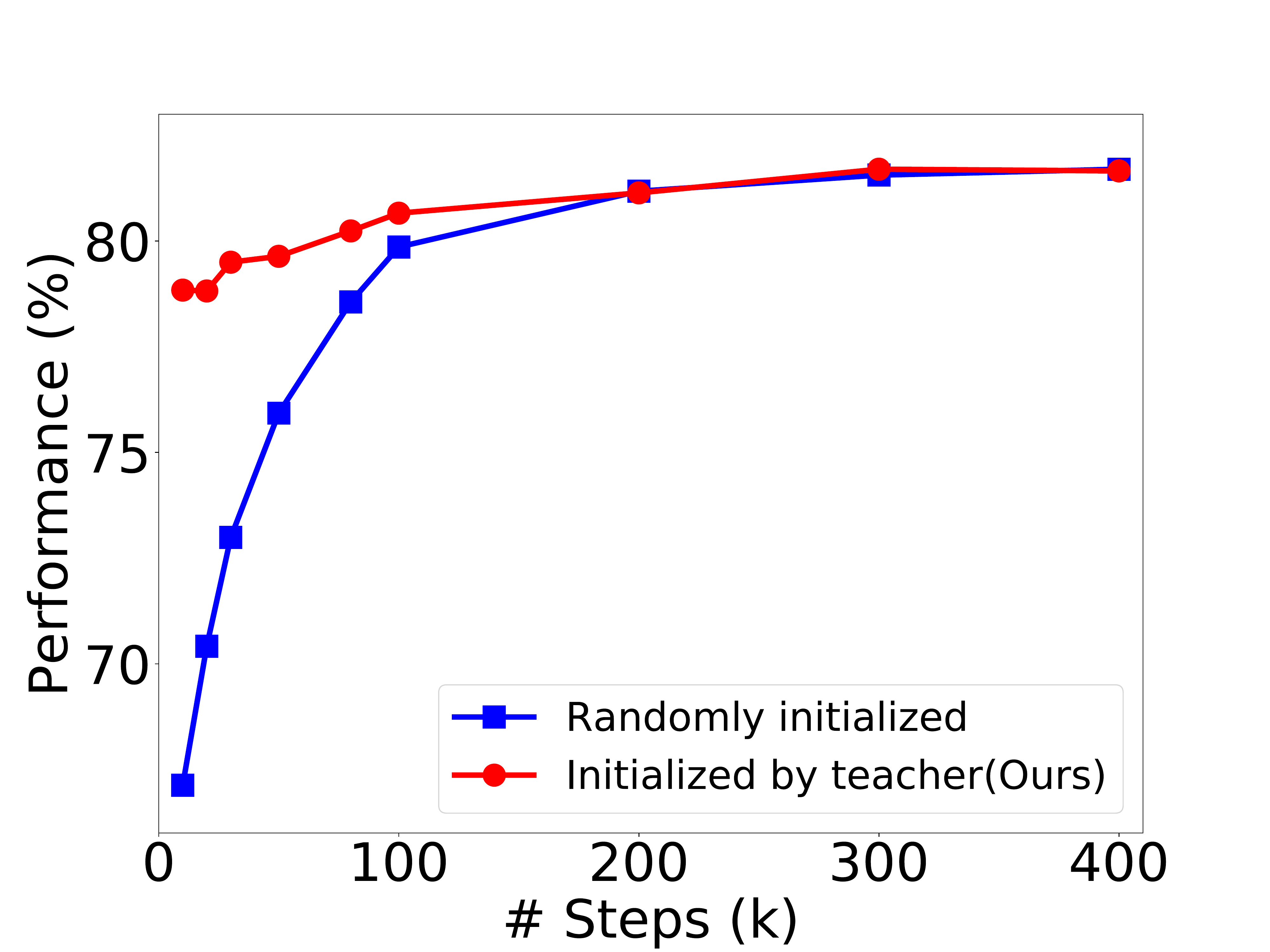}}
	\subfigure[Cross-Lingual]{\includegraphics[width=0.4\textwidth]{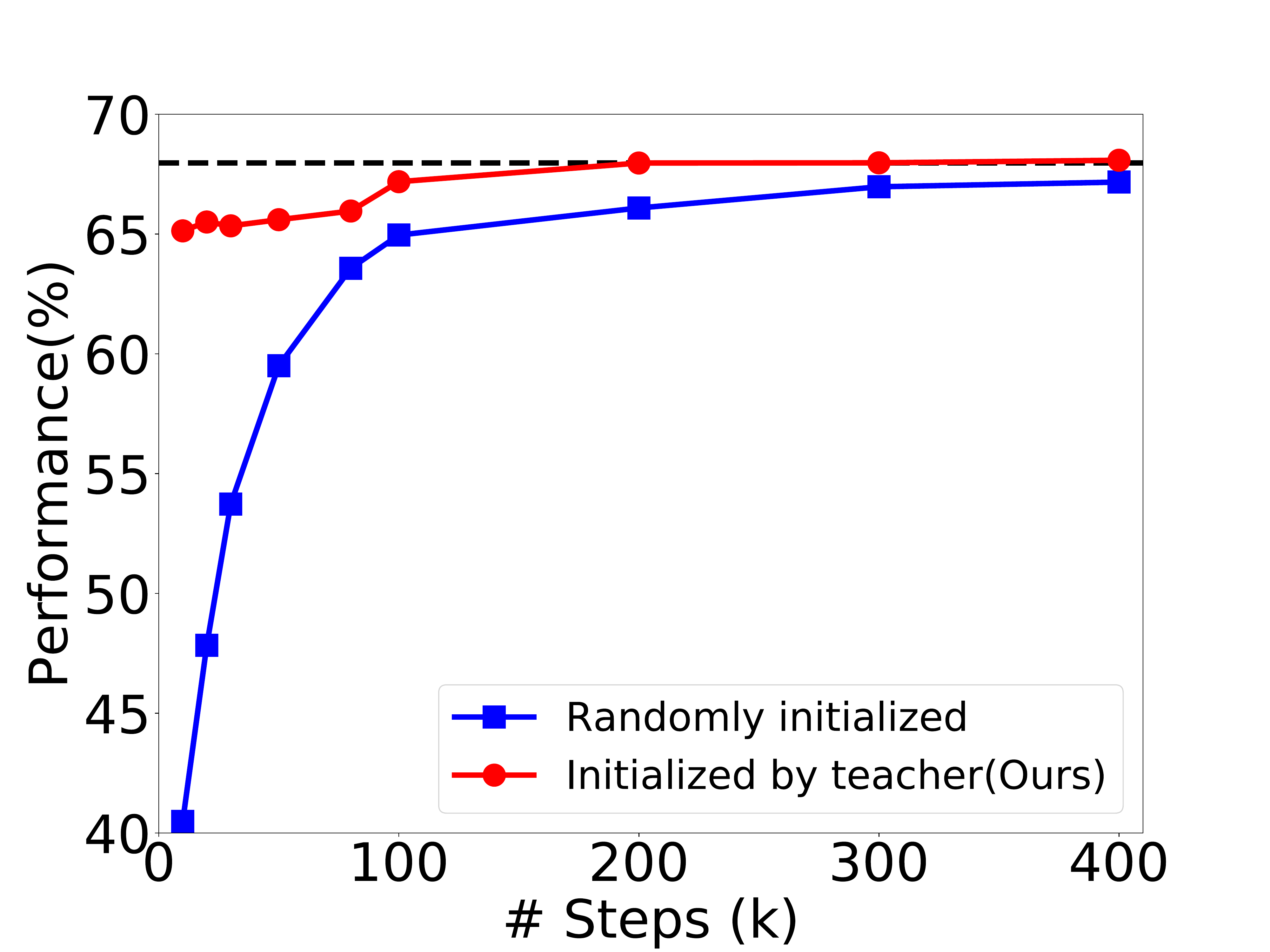}} 
	\vspace*{-0.4cm}
	\caption{The performance on English and other 5 languages~(cross-lingual) test sets with respect to the training steps. The horizontal dashed line in subfigure~(b) indicates the performance of our proposed initialization method at the step 200k.}
	\label{figure:initialization}
	\vspace{-0.2cm}
\end{figure}

\paragraph{Initialization} In Figure~\ref{figure:initialization}, we show the effects of initialization of the student model. Initialized by the embedding and bottom transformer layers of the teacher mBERT, our model can achieve a good result quickly on both English and other languages~(cross-lingual) test sets. Although random initialization has similar performance to our method on English test set after 200k training steps, it is still worse than ours on the cross-lingual test sets.    
%Specifically, at the training step of 200k, our model already outperforms the fully-trained randomly initialized student, which suggests initializing by our method can reduce the training time by a factor of 2. 

\begin{table}[H]
	\centering
	\begin{tabular}{|cccccccc|}
		\hline
		         \textbf{Model}           & \textbf{English} & \textbf{Spanish} & \textbf{Chinese} & \textbf{German} & \textbf{Arabic} & \textbf{Urdu} & \textbf{AVG} \\ \hline
		         MBERT(Teacher)           &       82.1       &       74.6       &       69.1       &      72.3       &      66.4       &     58.5      &     70.5     \\ \hline
		        LightMBERT(Ours)          &       81.5       &       74.7       &       69.3       &      72.2       &      65.0       &     59.3      &     70.3     \\
		\multicolumn{1}{|r}{-Freeze Emb.} &       80.9       &       74.2       &       68.0       &      70.8       &      65.6       &     58.9      &     69.7     \\ \hline
	\end{tabular}
	\vspace*{-0.2cm}
	\caption{Ablation study on the freezing the shared sub-word embeddings.}
	\label{tab:freeze}
\end{table}

\paragraph{Freezing the Embeddings} Our initial motivation of freezing the shared sub-word embeddings, initialized by that of the teacher mBERT, is to keep the sub-word embeddings of different languages in a shared space and ease the training process of the student. The results in Table~\ref{tab:freeze} show that freezing the shared embeddings druing the distillation process and fine-tuning stage can bring  0.6\% improvement. 

\begin{table}[H]
	\centering
	\begin{tabular}{|cccccccc|}
		\hline
		   \textbf{Model}     & \textbf{English} & \textbf{Spanish} & \textbf{Chinese} & \textbf{German} & \textbf{Arabic} & \textbf{Urdu} & \textbf{AVG} \\ \hline
		   MBERT(Teacher)     &       82.1       &       74.6       &       69.1       &      72.3       &      66.4       &     58.5      &     70.5     \\ \hline
		LightMBERT(Top-Layer) &       81.5       &       74.7       &       69.3       &      72.2       &      65.0       &     59.3      &     70.3     \\
		 LightMBERT(Uniform)  &       80.5       &       73.4       &       68.2       &      69.5       &      62.6       &     57.4      &     68.6     \\ \hline
	\end{tabular}
    \vspace*{-0.2cm}
	\caption{Comparison between the top-layer and uniform distillation strategy. Uniform indicates the student model learns evenly from the layers of the teacher mBERT.}
	\label{tab:layer_mapping}
\end{table}
\paragraph{Top-layer Distillation} We also conduct experiments to compare the top-layer distillation strategy with the uniform one that used in the original TinyBERT~\cite{jiao2019tinybert}. Specifically, we replace the top-layer distillation with the uniform one and keep other procedures in LightMBERT unchanged. The results in Table~\ref{tab:layer_mapping} show that the top-layer distillation strategy is better than the uniform one in the task-agnostic multilingual distillation setting, which is also observed in the recent monolingual KD work MiniLM~\cite{wang2020minilm} with a different distillation loss.

\section{Conclusion}
In this paper, we proposed a simple yet effective method to transfer the cross-lingual knowledge of a multilingual BERT to a light student model. Experiments empirically showed that our method significantly outperforms the baselines and achieves comparable results to the teacher mBERT. In addition, ablation studies confirmed the contribution of each procedure of the proposed method. In the future, we will evaluate our method on other cross-lingual tasks, such as NER, QA etc.

%\clearpage

\bibliographystyle{coling}
\bibliography{coling2020}

\end{document}